\documentclass[letterpaper, 10 pt, conference]{ieeeconf}  % Comment this line out if you need a4paper

\IEEEoverridecommandlockouts                              % This command is only needed if 
\usepackage{graphicx}
\usepackage{epsfig}

\usepackage{amssymb}
\usepackage{tabularx,colortbl}
\usepackage{array}
\usepackage{algorithmic}
\usepackage{algorithm}
\usepackage{graphicx}
\usepackage{epsfig}
\usepackage{hyperref}
\usepackage{tensor}
\usepackage{colortbl}
\usepackage{subcaption}
\usepackage{mwe}
\usepackage{color}
\graphicspath{{img/}}
\usepackage{bm}
\usepackage{url}
\usepackage{mathtools, nccmath}
\usepackage{graphicx}
\usepackage{epsfig}
\usepackage{bm}
\usepackage{mathtools}
\usepackage{algorithm,algorithmic}
\usepackage{graphicx}
\usepackage{subcaption}
\usepackage{hyperref}

\usepackage{hyperref}

\title{\LARGE \bf
Occlusion Handling by Pushing for Enhanced Fruit Detection
}
\author{Ege Gursoy$^{1,2}$ Dana Kulić$^{2}$ Andrea Cherubini$^{1}$% <-this % stops a space
\thanks{$^{1}$LIRMM, Univ. Montpellier, CNRS, Montpellier, France
        }%
\thanks{$^{2}$Monash University, Clayton, Australia
        }%
}

\begin{document}

\maketitle
\thispagestyle{empty}
\pagestyle{empty}

%%%%%%%%%%%%%%%%%%%%%%%%%%%%%%%%%%%%%%%%%%%%%%%%%%%%%%%%%%%%%%%%%%%%%%%%%%%%%%%%
\begin{abstract}
In agricultural robotics, effective observation and localization of fruits present challenges due to occlusions caused by other parts of the tree, such as branches and leaves. These occlusions can result in false fruit localization or impede the robot from picking the fruit. The objective of this work is to push away branches that block the fruit's view to increase their visibility. Our setup consists of an RGB-D camera and a robot arm. First, we detect the occluded fruit in the RGB image and estimate its occluded part via a deep learning generative model in the depth space. The direction to push to clear the occlusions is determined using classic image processing techniques. We then introduce a 3D extension of the 2D Hough transform to detect straight line segments in the point cloud. This extension helps detect tree branches and identify the one mainly responsible for the occlusion. Finally, we clear the occlusion by pushing the branch with the robot arm. Our method uses a combination of deep learning for fruit appearance estimation, classic image processing for push direction determination, and 3D Hough transform for branch detection. We validate our perception methods through real data under different lighting conditions and various types of fruits (i.e. apple, lemon, orange), achieving improved visibility and successful occlusion clearance. We demonstrate the practical application of our approach through a real robot branch pushing demonstration.
\end{abstract}

%%%%%%%%%%%%%%%%%%%%%%%%%%%%%%%%%%%%%%%%%%%%%%%%%%%%%%%%%%%%%%%%%%%%%%%%%%%%%%%%
\section{INTRODUCTION} \label{introduction}

In natural environments, fruits are often occluded due to the complex and dynamic nature of their surroundings. Occlusions occur when other parts of the tree, such as branches, leaves, or even neighboring fruits, cover the fruit. This phenomenon is prevalent in dense foliage where overlapping branches and leaves create complex obstructed sightlines. Occlusions lead to incomplete or inaccurate observations, causing inaccurate detection and localization results, and therefore decreasing the reliability of a fruit-picking robot. In general, all agricultural robot applications which rely on visual sensors (e.g., fruit harvesting, yield estimation, disease detection and phenotyping) would greatly benefit from more precise observations. In the literature of fruit detection and occlusion handling, existing approaches often focus on estimating the entire fruit's properties from partially occluded views. While these methods provide valuable insight, they inherently rely on estimations that may introduce uncertainties and inaccuracies. 

We propose a novel approach that goes beyond mere estimation. Rather than only relying on the prediction to find the appearance and location of occluded fruits, we take an    active step, by physically clearing the occlusions which obstruct the fruit from the robot view. Our setup consists of an RGB-D camera, for identifying occluded fruits and occluding branches, and a robot arm to clear the latter. This approach enables us to obtain more accurate fruit observations, thereby improving the precision and reliability of fruit detection in complex agricultural environments.

\begin{figure}[t!]
	\centering {\centering\includegraphics[width=0.99\columnwidth]{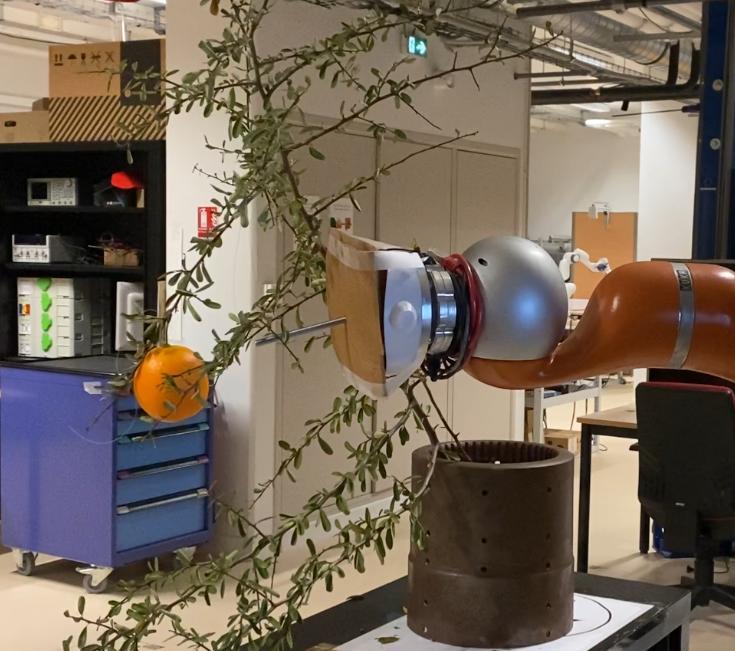}}
    \caption{The robot arm pushes an occluding branch to improve the view of an orange.}
    \label{fig:first} 
\end{figure}

To handle the occlusion by pushing the branch, we propose three steps: fruit identification, occlusion identification, and occlusion clearance. For fruit identification, we start by segmenting the visible section of the fruit in HSV color space, and then we apply a deep generative network to estimate the complete fruit in depth space. To identify occlusions, we design a guiding direction towards the occluding branch. Following this, we utilize HSV color space to segment branches attached to the tree. Then, we fit 3D line segments on the branches, via a 3D extension of the 2D Hough transform \cite{duda1972use}. To clear the occlusion, we analyze the branches within the view frustum of the fruit, and compare their orientation with the view towards the fruit. We select the branch closest to the view normal, which indicates the pushing direction. Then, we determine the push point farthest from the fruit within the view frustum to avoid robot self-occlusions. Finally, we control the robot to push away the occluding branch.

The main contributions of this work are:
\vspace{0.1cm}
\begin{itemize}
    \item we introduce a novel approach to handle occlusions in agricultural applications, by pushing branches away from the fruit view;
    \item we estimate the occluded parts of the fruit, by applying a deep generative network to the depth image;
    \item we extend Hough transform to 3D, to identify branches.
\end{itemize}
\vspace{0.1cm}

The rest of the paper is structured as follows. A review of the state of art in occlusion handling is given in Section \ref{sec:related_work}. Then, the problem statement and the overview of the method are explained in Sec.~\ref{sec:ourWork}. The fruit identification, occlusion identification and clearance are presented in Sections \ref{sec:fruit_identification}, \ref{sec:occlusion_identification} and \ref{sec:occlusion_clearance}, respectively. Finally, the results of the experiments are presented in Sec.~\ref{sec:experiment}, followed by a discussion and propositions for future work in Sec.~\ref{sec:conclusion}.

\section{RELATED WORK} \label{sec:related_work}

Detecting and locating fruits is an important challenge in agricultural robots. Detection-only deep learning algorithms have gained significant traction among researchers in the domain. In our previous work~\cite{gursoy-case23}, deep learning algorithms were used to detect and track oranges in images acquired by the robot RGB-D camera. Santos et al. \cite{santos2020grape} introduced an interactive instance mask generation method on RGB images, comparing the performance of various learning-based object detection algorithms on grape detection. Kang et al. \cite{kang2020fast} developed an automatic label generation module tailored for detecting apples in RGB images of orchards. Similarly, Koirala et al. \cite{koirala2019deep} investigated popular object detection frameworks for mango detection, devising a custom detection network. Liu et al. \cite{liu2018visual} adopted a multi-class approach for citrus detection by categorizing samples into distinct classes. Kirk et al. \cite{kirk2020b} implemented a custom neural network that incorporated a 6-channel RGB + CIELAB representation as input for fruit detection tasks. Additionally, Ganesh et al. \cite{ganesh2019deep} trained a neural network using RGB and HSV 6-channel input for orange detection. Despite the effectiveness of these deep learning approaches, they are complex and not robust to occlusions. This emphasizes the ongoing need for adaptable detection methodologies.

Detection-only methods solely focus on improving the accuracy and efficiency of fruit detection, without addressing occlusions. Gao et al. \cite{gao2020multi} categorized apples based on occlusion conditions, to facilitate subsequent picking, while Fu et al. \cite{fu2018kiwifruit} categorized kiwifruits based on distance. Deep learning-based detectors have gained popularity, with approaches ranging from instance mask generation to custom neural networks, tailored for specific fruits.

Various approaches have been proposed to estimate occlusions on fruit. Feng et al. \cite{feng2015design} developed a tomato-picking robot utilizing structured light and color thresholding. However, occlusion remained a significant cause of picking failures, as highlighted in Ling et al. \cite{ling2019dual} for a tomatoes use case. Kurtulmus et al. \cite{kurtulmus2014immature} employed statistical classifiers and neural networks to detect peaches, although challenges persisted with high occlusions. These methods indicate a growing focus on completing occluded fruits to enhance efficiency and reliability.

Branch detection methods also play a crucial role. Chen et al. \cite{chen2015reasoning} presented the dense composition of tomato plants, often occluded by branches, emphasizing the importance of accurate branch detection for effective fruit harvesting. Gong et al. \cite{gong2022robotic} introduced a high-precision  depth learning method for estimating occluded targets, while Lv et al. \cite{lv2016recognition} developed a method for identifying occluded fruits in natural environments, utilizing dynamic threshold segmentation. These studies demonstrate the significance of branch detection, for overcoming occlusions in agrirobot applications.

In addition to technological solutions, Li et al. \cite{li2022novel} experimented removing leaves beneath plants, to reduce target occlusion, while Gené-Mola et al. \cite{gene2020fruit} applied airflow using sprayers to diminish occlusions. These labor intensive methods, demonstrate the potential of agronomic interventions in improving the visibility and accessibility of fruits.

The state-of-the-art primarily focuses on improving estimations. Our work distinguishes itself by prioritizing the clearing of occlusions, i.e. identifying the ground truth rather than relying solely on estimations.

\section{OUTLINE OF OUR WORK} \label{sec:ourWork}
\subsection{Problem Statement} \label{sec:problem}

\begin{figure}[t!]
  \begin{subfigure}{.47\columnwidth}
  \centering
    \includegraphics[width=.95\linewidth]{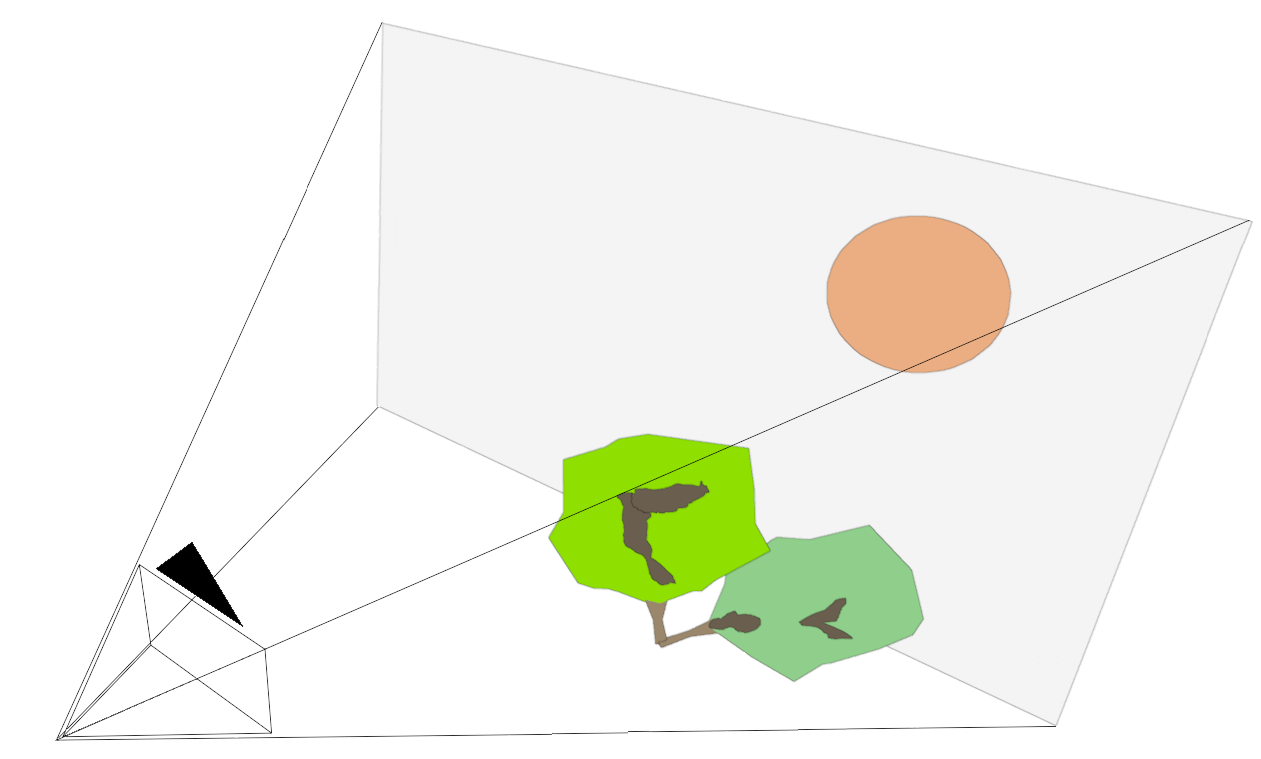}
  \end{subfigure}%
  \begin{subfigure}{.47\columnwidth}
  \centering
    \includegraphics[width=.95\linewidth]{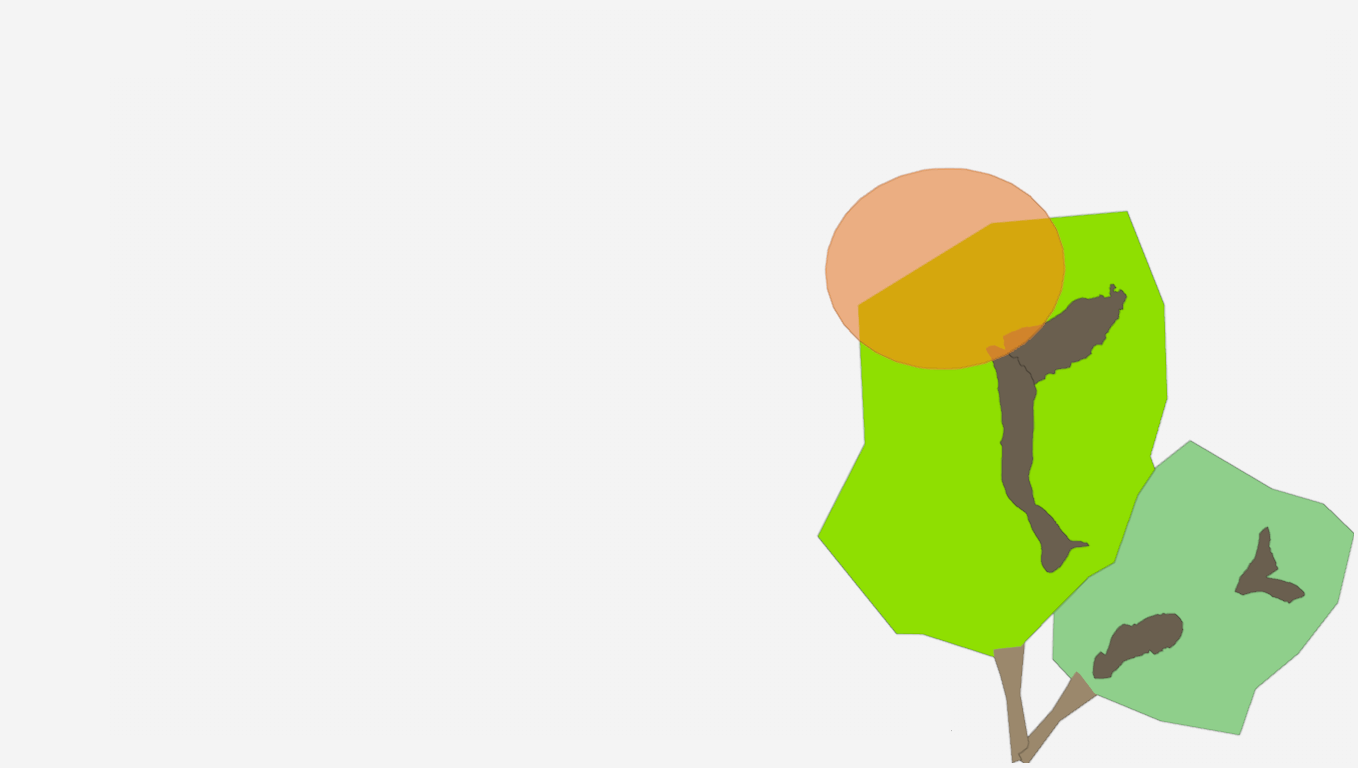}
  \end{subfigure}
  \caption{Camera field of view (left), with corresponding image (right). The tree branches and leaves are green/brown, and the fruit is orange.}
  \label{fig:fov}
\end{figure}

Consider a robot observing with a RGB-D camera a fruit tree. The goal of our work is to push, with the robot arm, the occlusions, which limit the visibility of the fruit. Fruits are partially observed by the robot because of the occlusions caused by other tree parts such as branches and leaves (see Figure \ref{fig:fov}). These occlusions lead to two significant problems. Firstly, the precision of fruit localization is compromised, since the robot can only perceive a partial view of the fruit. Second, creating a clear path for either the second arm or a person to access the fruit. Agricultural settings inherently contain numerous complex elements, and detectors may incorrectly identify fruits in shaded areas, leading to instances where a detected object is mistakenly identified as a fruit. These challenges emphasize the importance of addressing occlusions for more accurate and reliable fruit detection and localization. 
%To address these issues, we propose a method for identifying partially observed fruits and estimating the complete fruit in Section \ref{sec:fruit_identification}. Then, in Section \ref{sec:occlusion_identification} we identify the pushing direction that could decrease the occlusion, and find the related branch to push (Sec.~\ref{sec:occlusion_clearance}).

We define two hypothesis in this work:
\begin{itemize}
    \item \textit{Hypothesis 1}: Undetected pixels of the fruit are in the camera field of view, reducing the image dimensions.
    \item \textit{Hypothesis 2}: The fruits in this work are considered perfectly elliptical in shape, simplifying the complexity of the calculations.
\end{itemize}

% \begin{figure}[t!]
% \centering

% \begin{subfigure}{0.5\columnwidth}
%   \centering
%   \includegraphics[width=0.8\linewidth]{overview.png}
%   \caption{}
%   \label{subfig:fig1}
% \end{subfigure}

% \begin{subfigure}{0.5\columnwidth}
%   \centering
%   \includegraphics[width=0.8\linewidth]{cam_raw.png}
%   \caption{}
%   \label{subfig:fig2}
% \end{subfigure}

% \caption{Camera field of view (a), with corresponding image (b). The tree branches and leaves are green/brown, and the fruit is orange.}
% \label{fig:fov}
% \vspace{-0.6cm}
% \end{figure}

\subsection{Overview of our method} \label{sec:ow_vision}

Our approach begins by segmenting the visible portion of the fruit using color information, to isolate the visible pixels of the fruit. These pixels are projected on the depth image, to obtain the corresponding 3D points. Then, we use a generative network to estimate the complete fruit in 2D. To this end, we only focus on the depth image, to reduce the network complexity. The network is trained using pairs of occluded/complete fruit depth images; it takes an occluded fruit depth image as input and generates an estimation of the complete fruit depth image. We determine the direction vector from the centroid of the estimated fruit to the centroid of the difference between the segmented and estimated fruit to calculate the direction needed to clear the occlusion (named \textit{view gradient}). Further, we compute the imaginary line from the camera to the centroid of the estimated fruit, and we create the \textit{view frustum} of the fruit around that line. We explore the point cloud of the environment generated by the RGB-D camera, and identify the points within the view frustum as the occlusion points. We search for 3D line segments in the point cloud and select the \textit{pushing line}, based on their relationship with the view gradient and on their distance to the view frustum. Then, we define the \textit{pushing point} depending on the location of the pushing line wrt the view frustum. The amount and direction of the push is set according to the view gradient and to the fruit size. Finally, we push the point with the robot arm, to clear the occlusion.

% The amount of the push is set proportionally to the size of the occlusion and adjusted based on the empty spaces in the point cloud opposite the view gradient. This adjustment is necessary to prevent the generation of additional obstructions during the pushing operation. TODO.

\section{FRUIT IDENTIFICATION} \label{sec:fruit_identification}

\subsection{Partial Fruit Segmentation} \label{sec:partial_fruit_segmentation}

Let us denote $\mathbf{m}$, the mapping between pixel $\left(u,v\right) \in \mathcal{I} = \left\{1, \dots, w\right\} \times \left\{1, \dots, h\right\} \subset \mathcal{N}^2$ (from a camera of $w \times h$ resolution) and the 3D position of the corresponding point in the camera frame $\mathbf{P} = (x,y,z) \in \mathcal{F}_{cam} \subset \mathcal{R}^{3}$. We denote the projection of a pixel $(u,v)$ in color information towards the depth information as $z_{uv} \in \mathcal{R}^+$. If we consider an un-distorted perspective projection (i.e., a pinhole model), the mapping is:
\begin{equation} 
\begin{array}{rccc}
\mathbf{m}:  & \mathcal{I} \times \mathcal{R}^+  & \rightarrow & \mathcal{F}_{cam} \vspace{.2cm}\\
& \left( 
\begin{array}{c}
u\\
v\\
z_{uv}
\end{array}
\right) & \mapsto & 
\left( 
\begin{array}{c}
x=\frac{u-u_0}{f} z_{uv}\\
y=\frac{v-v_0}{f} z_{uv}\\
z=z_{uv}
\end{array}
\right)
\end{array}
\label{eq:proj}
\end{equation}
with $(u_0,v_0) \in \mathcal{I}$ the camera principal point coordinates in the image plane and $f \in \mathcal{R}^+$, the camera focal length. Note that  this mapping is \textit{bijective}: 
\begin{equation}
\begin{array}{rccc}
\mathbf{m}^{-1}:  & \mathcal{F}_{cam}  & \rightarrow &  \mathcal{I} \times \mathcal{R}^+ \vspace{.2cm}\\
& \left( 
\begin{array}{c}
x\\
y\\
z 
\end{array}
\right) & \mapsto & 
\left( 
\begin{array}{c}
u=xf / z_{uv} + u_0\\
v=yf / z_{uv} + v_0\\
z=z_{uv}
\end{array}
\right)
\end{array}
\label{eq:proj-1}
\end{equation}

To find a fruit, we analyze the color image and specifically its Hue (H), Saturation (S), Value(V) histograms. We determine the H, S, V values corresponding to fruits and segment the corresponding pixels. We then retrieve the $z$ values of these pixels from the depth, forming the depth image of the segmented fruit $\mathbf{D}_f$.

% We apply \ref{eq:proj} to  $\left(\mathbf{u}_{s},\mathbf{v}_{s},\mathbf{z}_{s}\right)^T \in \mathcal{I}^a \times \mathcal{R}^{a}$ where $a$ number of segmented pixels which belong to the fruit, to get $\mathbf{P}_s = ({\mathbf{x}_s,\mathbf{y}_s,\mathbf{z}_s})^T\in \mathcal{R}^{3\times a}$, their corresponding 3D positions. We eliminate outlying points based on their Z-Score by:

% We apply \ref{eq:proj} to  $\left(\mathbf{u}_{s},\mathbf{v}_{s},\mathbf{z}_{s}\right)^T \in \mathcal{N}^{w\times a} \times \mathcal{N}^{h\times a} \times \mathcal{R}^{a}$ where $a$ number of segmented pixels which belong to the fruit, to get $\mathbf{P}_s = ({\mathbf{x}_s,\mathbf{y}_s,\mathbf{z}_s})^T\in \mathcal{R}^{3\times a}$, their corresponding 3D positions. We eliminate outlying points based on their Z-Score by the equation \ref{eq:zscore}.

% \begin{equation} \label{eq:zscore}
% \mathbf{z}_{score} = ({{\mathbf{P}_s} - \bm{\mu}}) \oslash {\bm{\sigma}}
% \end{equation}

% with $\bm{\mu}, \bm{\sigma} \in \mathcal{R}^{3}$ are the mean and the standard deviation of the 3D points in each axes, and $\oslash$ as the Hadamard division. 
% Points $\mathbf{P}_s$ whose $\mathbf{z}_{score}$ exceeds a defined threshold on any axis, are considered outliers and not part of the fruit. The remaining inlier points $i = a-o$ points with $o$ number of outliers, form $\mathbf{P}_f = (\mathbf{x}_f,\mathbf{y}_f,\mathbf{z}_f)^T \in \mathcal{R}^{3 \times i}$, 3D points belong to the fruit.

\subsection{Fruit Appearance Estimation} \label{sec:fruit_estimation}

\begin{figure}[t!]
	\centering 
	\includegraphics[width=0.99\columnwidth]{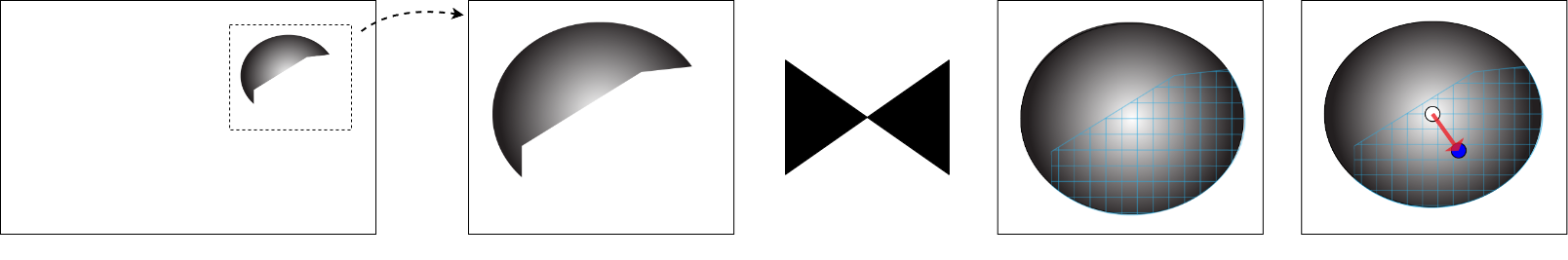}
	\caption{Outline of the fruit estimation method. Left to right: Visible fruit pixels in masked depth image $\mathbf{D}_M$, cropped ROI depth image $\mathbf{D}'_M$ fed into generative model which outputs the estimated fruit depth image, center point of the fruit (white circle) $(\hat{u}^\prime_c, \hat{v}^\prime_c)$, center point of the $\mathbf{\bar{M}}^\prime$ (blue circle) defined as $(\bar{u}^\prime_c, \bar{v}^\prime_c)$ shown on $\mathbf{\bar{D}}_M^\prime$, occlusion direction vector $\vec{\mathbf{{O}}}_f$ (red arrow). Cross hatched area represents the area $\mathbf{\hat{D}}_M^\prime$, estimated by the generative model.}
 \label{fig:crop_estimation}
\end{figure}

From $\mathbf{D}_f$, we estimate the occluded part of the fruit using a generative neural network, based on an encoder-decoder architecture. Figure~\ref{fig:crop_estimation} outlines the method.

First, we apply a 2D binary mask $\mathbf{M} \in \mathcal{I} \times \{0,1\}$ to $\mathbf{D}_f$. The resulting masked depth image is:
\begin{equation}
    \label{eq:get_maskedD}
    \mathbf{D}_{M} = \mathbf{M} \otimes \mathbf{D}_f \in \mathcal{I} \times \mathcal{R}^+,
\end{equation}
with $\otimes$ as the Hadamard product. A visual representation of $\mathbf{D}_{M}$ is shown in the leftmost picture of Figure \ref{fig:crop_estimation}.

Under \textit{Hypothesis 1} defined in Section \ref{sec:problem}, the complete fruit should appear (if there were no occlusions) inside a rectangular region of interest (ROI). 

% We find the centroid $(u_b,v_b)$ of the detected blob in $\mathbf{B}_f$ by:

% We locate the point with minimum $z$ in the segmented fruit depth image $\mathbf{D}_{f}$, defined as  $(x_b,y_b,z_b)^T $. 

We crop $\mathbf{D_M}$ to obtain the ROI $\mathbf{D}^\prime_M$ by (\ref{eq:crop}):

% \begin{equation}
%     \begin{aligned}
%         z_{b} = \min\{z \mid \forall (x, y, z) \in \mathbf{P}_{f}\}\\
%         (x_{b}, y_{b}) = (x_i, y_i), \mathrm{where}~~ z_i = z_{b}
%     \end{aligned}
% \end{equation}

\begin{equation} 
\begin{aligned}
\mathbf{D}_{M}^\prime = \{ \mathbf{D}_{M} \mid u_{TL} \leq u_i \leq u_{BR}, v_{TL} \leq v_j \leq v_{BR}\}\\
\mathrm{with}~~  (u_{TL}, v_{TL})=w_s - d_{ROI}/2, h_s - d_{ROI}/2\\
     (u_{BR}, v_{BR})=w_s + d_{ROI}/2, h_s + d_{ROI}/2
\end{aligned}
\label{eq:crop}
\end{equation}
with $w_s$, $h_s$ width and height of the segmented fruit and $d_{ROI}$ ROI size, $u_{TL},v_{TL}$ and $u_{BR},v_{BR}$ as the top-left and bottom-right pixels of the ROI.

It is always possible to reinsert the ROI in the original image, by applying translation $\mathbf{r}$:

\begin{equation} 
\begin{array}{rccc}
\mathbf{r}:  & \mathbf{D}_{M}^\prime  & \rightarrow & \mathbf{D}_{M} \vspace{.2cm}\\
& \left( 
\begin{array}{c}
u\\
v
\end{array}
\right) & \mapsto & 
\left( 
\begin{array}{c}
u^\prime = u + u_{TL}\\
v^\prime = v + v_{TL}
\end{array}
\right).
\end{array}
\label{eq:mappingR}
\end{equation}

The network is trained with pairs of occluded and non-occluded $\mathbf{D}_M^\prime$. Both the fruits and occlusions are generated at random locations. Non-occluded orange images are synthetic, created from ellipses colored similarly to the fruit, and occlusions are simulated by random croppings on the non-occluded images. The dataset consists of 100 images, divided 70:30 into training and validation sets. We feed the input with the occluded depth image and construct the loss by comparing the generated image and the non-occluded image.

The encoder-decoder network uses a modified U-net image generator from our previous work \cite{gursoy2023tro}, with the following specifications. For the encoder, the size of the input layer is fixed to 56x56. The input is followed by 4 downscale layers, each consisting of two 3x3 convolution layers followed by a 2x2 maxpooling layer. Filter sizes of the convolution layers are 32, 64, 128 and 256, respectively. The decoder has 5 upscale layers and 1 output layer. Upscale layers consist of two 3x3 convolution layers, followed by a 2x2 upsampling layer. Filter size of convolution layers are 512, 256, 128, 64 and 32, respectively. Outputs of the second convolution layer from downscale layers are concatenated with the output of the symmetrical upsampling layer, before being fed to the following upscale layers. The output layer is a 1x1 convolution layer with sigmoid activation function. All the strides are set to 1 and rectified linear unit activation is used in all the convolution layers, except for the output layer. We use the binary cross entropy loss function in our network.

After training, giving an occluded $\mathbf{D}_M^\prime$ as input, we can generate predicted image $\mathbf{\hat{D}_M^\prime}$ of the completed ROI. We can obtain the centroid pixels $\hat{u}^\prime_c$, $\hat{u}^\prime_c$ in $\mathbf{\hat{D}}_M^\prime$ via:
\begin{equation} \label{eq:centroid}
\hat{u}^\prime_c, \hat{v}^\prime_c = \frac{1}{N_{\neq 0}} \sum_{i,j | z_{i,j} \neq 0} u_{i,j}, ~~\frac{1}{N_{\neq 0}} \sum_{i,j | z_{i,j} \neq 0} v_{i,j}
\end{equation}
with $N_{\neq 0}$ the number of fruit pixels with $z_{i,j} \neq 0$ and $u_{i,j}$, $v_{i,j}$ their pixel coordinates. We also compute $\hat{w}_f, \hat{h}_f$, i.e., the width and height of the estimated fruit by subtracting maximum and minimum pixels with $z_{i,j} \neq 0$.

% \begin{equation}
% \begin{aligned}
%     \hat{w}_f, \hat{h}_f &= \hat{u}_{max} - \hat{u}_{min}, \hat{v}_{max} - \hat{v}_{min}\\
%     \mathrm{s.t.}~~ &\hat{u}_{max} = \arg \max_{i,j}\{u_{i,j} \mid z_{i,j} \neq 0\}\\
%                     &\hat{v}_{max} = \arg \max_{i,j}\{v_{i,j} \mid z_{i,j} \neq 0\}\\
%                     &\hat{u}_{min} = \arg \min_{i,j}\{u_{i,j} \mid z_{i,j} \neq 0\}\\
%                     &\hat{v}_{min} = \arg \min_{i,j}\{v_{i,j} \mid z_{i,j} \neq 0\}
% \end{aligned}
% \end{equation}

Finally, we calculate the projection of $(\hat{u}^\prime_c, \hat{v}^\prime_c)$ in $\mathcal{F}_{cam}$:
\begin{equation}
\hat{\mathbf{p}}_c = 
\left( 
\begin{array}{c}
\hat{x}_c\\
\hat{y}_c\\
\hat{z}_c
\end{array}
\right)
=
\mathbf{m}\left( 
\begin{array}{c}
\hat{u}_c\\
\hat{v}_c\\
z_{\hat{u}_c,\hat{v}_c}
\end{array}
\right)
\label{eq:point_centroid}
\end{equation}
% with $\mathrm{dist}$ as the distance between two points in 3D Cartesian space.
% TODO: This should be changed and calculated from the prediction, not detection.
with $\hat{u}_c, \hat{v}_c = \mathbf{r}[(\hat{u}^\prime_c, \hat{v}^\prime_c)]$.

In the rest of the paper, we consider the properties of the real (occluded) fruit identical to those of its estimated (entire) counterpart; i.e., we assume that the real fruits' properties ${p}_c, {w}_f, {h}_f, {u}_c, {v}_c$  are identical to estimated $\hat{p}_c, \hat{w}_f, \hat{h}_f, \hat{u}_c, \hat{v}_c$ and the two sets can be used interchangeably.

\section{OCCLUSION IDENTIFICATION} \label{sec:occlusion_identification}

\subsection{View Gradient}  \label{sec:view_graident}

We generate two binary mask images $\mathbf{{M}^\prime}$ and $\mathbf{\hat{M}^\prime}$ from $\mathbf{{D}}_M^\prime$ and $\mathbf{\hat{D}}_M^\prime$, wherein pixels are set to 1 if their values differ from 0. We define $\mathbf{\bar{M}}^\prime = |\mathbf{M}^\prime - \mathbf{\hat{M}}^\prime|$. We compute $area(\mathbf{\bar{M}}^\prime)/area(\mathbf{M}^\prime)$. If this ratio is bigger than a threshold $r_o$, we consider that the fruit has an occlusion. In this case, we compute $(\bar{u}^\prime_c, \bar{v}^\prime_c)$ the centroid of $\mathbf{\bar{M}}^\prime$ by adapting (\ref{eq:centroid}) and we calculate the occlusion pixels direction (i.e., unitary vector) $\vec{\mathbf{O}}_f \in \mathcal{R}^{2}$:
\begin{equation}
\vec{\mathbf{O}}_f = \frac{\left(\bar{u}^\prime_c - \hat{u}^\prime_c, \bar{v}^\prime_c - \hat{v}^\prime_c\right)^T}{\lVert\bar{u}^\prime_c - \hat{u}^\prime_c, \bar{v}^\prime_c - \hat{v}^\prime_c\rVert}.
\end{equation}
This vector, named \textit{view gradient}, determines the direction of the most missings, and is the red arrow shown in Figures \ref{fig:crop_estimation} and \ref{fig:occlusionid}.

\subsection{Branch Line Segments}

First, we compute the H, S, V histograms as in \ref{sec:partial_fruit_segmentation}. Then, we determine the branches values, to segment the branch pixels. We binarize the image by applying a threshold: if $z > 0$, the pixel is set to 1; otherwise, it is set to 0. We apply a Hough transform, to find straight line segments \cite{duda1972use} $\mathbf{l}_{2D}$ defined by their image end points: $(u_{l1}, v_{l1})$ and $ (u_{l2}, v_{l2})$. The main advantages of the Hough transform over other line fitting methods are its simultaneous detection of all lines. We design an extension of the Hough transform in 3D space to fit line segments $\mathbf{l}_{branch}$ into those branch points. 

To this end, we define a distance threshold ${\rho}_{th}$ and an angle threshold $\theta_{th}$. We compare all detected lines and remove similar lines if their Hough distances wrt $\rho$ and $\theta$ are smaller than ${\rho}_{th}$ and $\theta_{th}$. For each line segment $\mathbf{l}_{2D}$, we determine the plane $\mathbf{\Pi}$ in $\mathcal{F}_{cam}$, passing through the camera center and the line segment. The endpoints of the line segment in the image are projected onto the 3D camera frame $\mathcal{F}_{cam}$ using the mapping $\mathbf{m}$, where their $z$ coordinate is set to focal length $f$. Plane $\mathbf{\Pi}$ can be expressed as:
\begin{equation}
Ax+By+Cz = 0, 
\end{equation}
with $\mathbf{N}_{\Pi} = (A,B,C) = (u_{l1}, v_{l1}) \times (u_{l2}, v_{l2})$ its normal vector.

To extend the Hough transformation to 3D, we project the branch pixels into depth information and form $\mathbf{P}_b$, branch points using mapping $\mathbf{m}$. Points in $\mathbf{P}b$ that are closer than a distance threshold $d_{\Pi}$ to the plane $\mathbf{\Pi}$ are filtered out, resulting in the formation of $\mathbf{P}^{\Pi}_b$. Then, we rotate the points in $\mathbf{P}^{\Pi}_b$ around the camera center point to align the plane $\mathbf{\Pi}$ with the camera optical axis to form $'\mathbf{P}^{\Pi}_b$. The rotation matrix $\mathbf{R}$ about an arbitrary axis by angle $\alpha$ is calculated by Rodrigues' formula (with $\mathcal{F}^{z}_{cam}$ denoting the $z$ axis of $\mathcal{F}_{cam}$):
\begin{equation}
\label{eq:rodriguez}
\begin{aligned}
    \mathbf{K} &= \mathcal{F}^{z}_{cam} \times \mathbf{N}_{\Pi} \\
    \cos(\alpha) &= \frac{\mathcal{F}^{z}_{cam} \cdot \mathbf{N}_{\Pi}}{\mathcal{F}^{z}_{cam} \cdot \mathbf{N}_{\Pi}}\\
    '\mathbf{P}^{\Pi}_b &= \mathbf{R}(\mathbf{K},\alpha) \cdot \mathbf{P}^{\Pi}_b.
\end{aligned}
% \vspace{-0.5cm}
\end{equation}

\begin{figure}[t!]
	\centering {\centering\includegraphics[width=0.75\columnwidth]{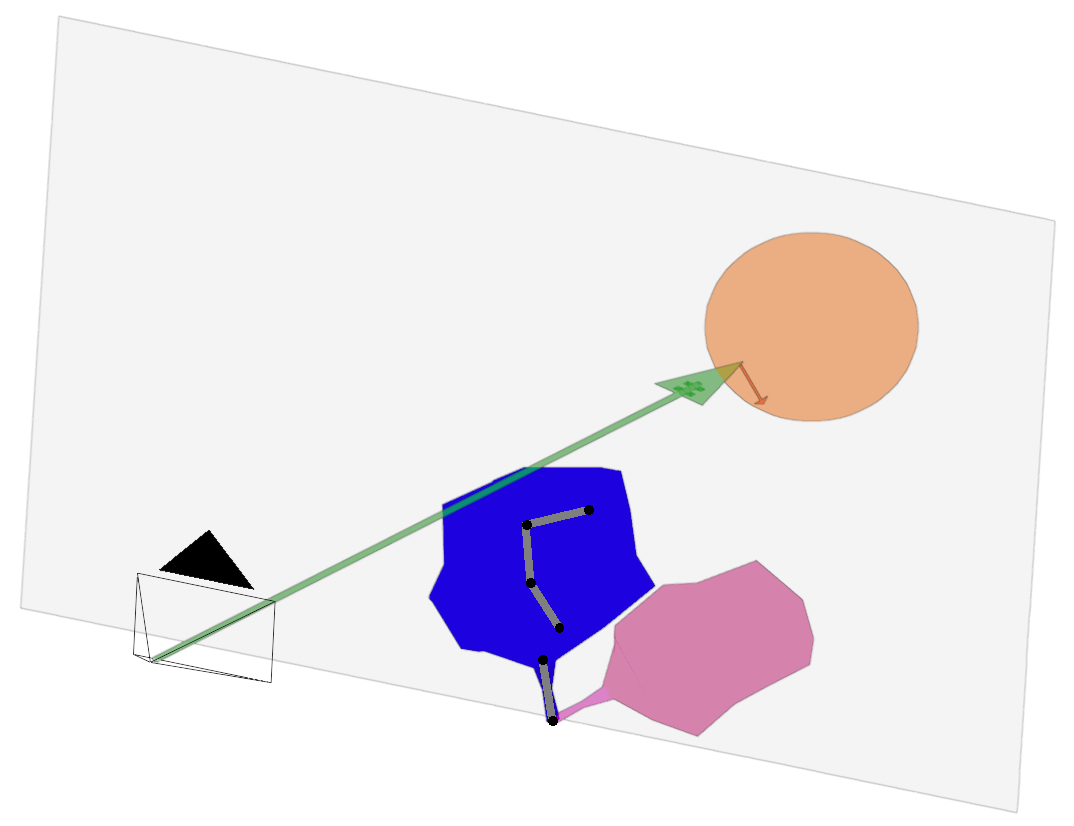}}
    \caption{Line segments (grey) and their extremities (black circles). The pushing line $\mathbf{l}_{push}$ $\mathbf{l}_{push}$ is selected as the topmost according to Algorithm \ref{alg:line_decision}.}
    \label{fig:hough} 
\end{figure}

We form a new 2D image from $'\mathbf{P}^{\Pi}_b$ and reapply Hough transform to find straight line segments and remove similar lines to get $\mathbf{l}^{'}_2D$. Since the line detected by the last process is an intersection of two planes, the resulting line can be expressed as a 3D line: $\mathbf{l}^{'}_3D = \mathbf{m}(\mathbf{l}^{'}_2D)$. We apply the inverse of the applied rotation to finally obtain $\mathbf{l}_{3D} = \mathbf{R}(\mathbf{K},-\alpha) \cdot \mathbf{l}^{'}_{3D}$, the 3D line segment in $\mathcal{F}_{cam}$. We define $\mathbf{l}_b$, containing all the $\mathbf{l}_{3D}$ from all of the $\mathbf{l}_{2D}$. 

All line finding steps are described in algorithm \ref{alg:line_find}, where the image processing and line segment post-processing steps are shortened by \textit{2D Hough line transform}.

\begin{algorithm}
 \caption{Line finding algorithm}
 \begin{algorithmic}[1]
 \renewcommand{\algorithmicrequire}{\textbf{Input:}}
 \renewcommand{\algorithmicensure}{\textbf{Output:}}
 \REQUIRE Color and Depth information
 \ENSURE 3D branch line segments $\mathbf{l}_{b}$
 \STATE Branch segmentation in HSV
 \STATE Form an image from segmented pixels
 \STATE 2D Hough line transform on the image: $\mathbf{l}_{2D}$
    \FOR{$\mathbf{l} \in \mathbf{l}_{2D}$ }
        \STATE Compute the plane $\Pi$
        \STATE Filter $\mathbf{P}_b$ by distance to $\Pi$: $\mathbf{P}^{\Pi}_b \leftarrow \mathbf{P}_b < d_{\Pi}$
        \STATE Rotation of the points $\mathbf{P}^{\Pi}_b$: $'\mathbf{P}^{\Pi}_b \leftarrow \mathbf{R}(\mathbf{K},\alpha) \cdot \mathbf{P}^{\Pi}_b$
        \STATE Form a new image from  $'\mathbf{P}^{\Pi}_b$:
        \STATE 2D Hough transform on the new image: $\mathbf{l}^{'}_{2D}$
        \STATE 3D mapping of $\mathbf{l}^{'}_{2D}$: $\mathbf{l}^{'}_{3D} \leftarrow \mathbf{m}(\mathbf{l}^{'}_{2D})$
        \STATE Inverse rotation of $\mathbf{l}^{'}_{3D}$: $\mathbf{l}_{3D} \leftarrow \mathbf{R}(\mathbf{K},-\alpha) \cdot \mathbf{l}^{'}_{3D}$
    \ENDFOR
    \RETURN $\mathbf{l}_b$ which contains all the $\mathbf{l}_{3D}$
 \end{algorithmic} 
\label{alg:line_find}
\end{algorithm}

\section{OCCLUSION CLEARANCE}\label{sec:occlusion_clearance}

\subsection{Line Decision}

% \begin{figure}[t!]
% 	\centering {\centering\includegraphics[width=0.9\columnwidth]{front2.png}}
%     \caption{Orthogonal projection of the environment from the camera to the image center. Yellow and orange represent the view cylinder and the fruit. The base of the view cylinder is slightly bigger than the fruit as a result of the \textit{Hypothesis 3}.}
%     \label{fig:cylinder}
% \end{figure}

\begin{figure}[t!]
\centering

\begin{subfigure}{0.5\columnwidth}
  \centering
  \includegraphics[width=0.9\linewidth]{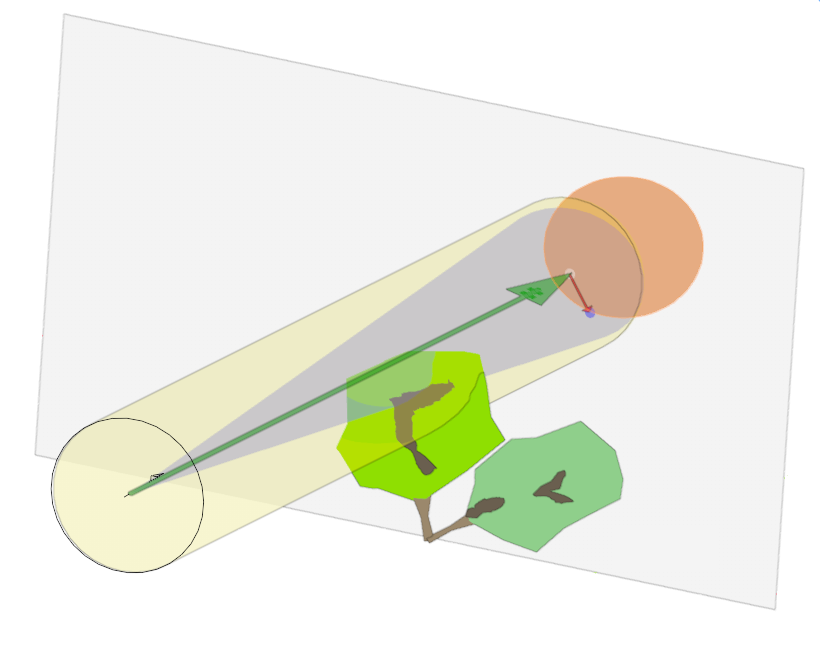}
  \label{subfig:fig1}
\end{subfigure}%
\hfill
\begin{subfigure}{0.5\columnwidth}
  \centering
  \includegraphics[width=0.9\linewidth]{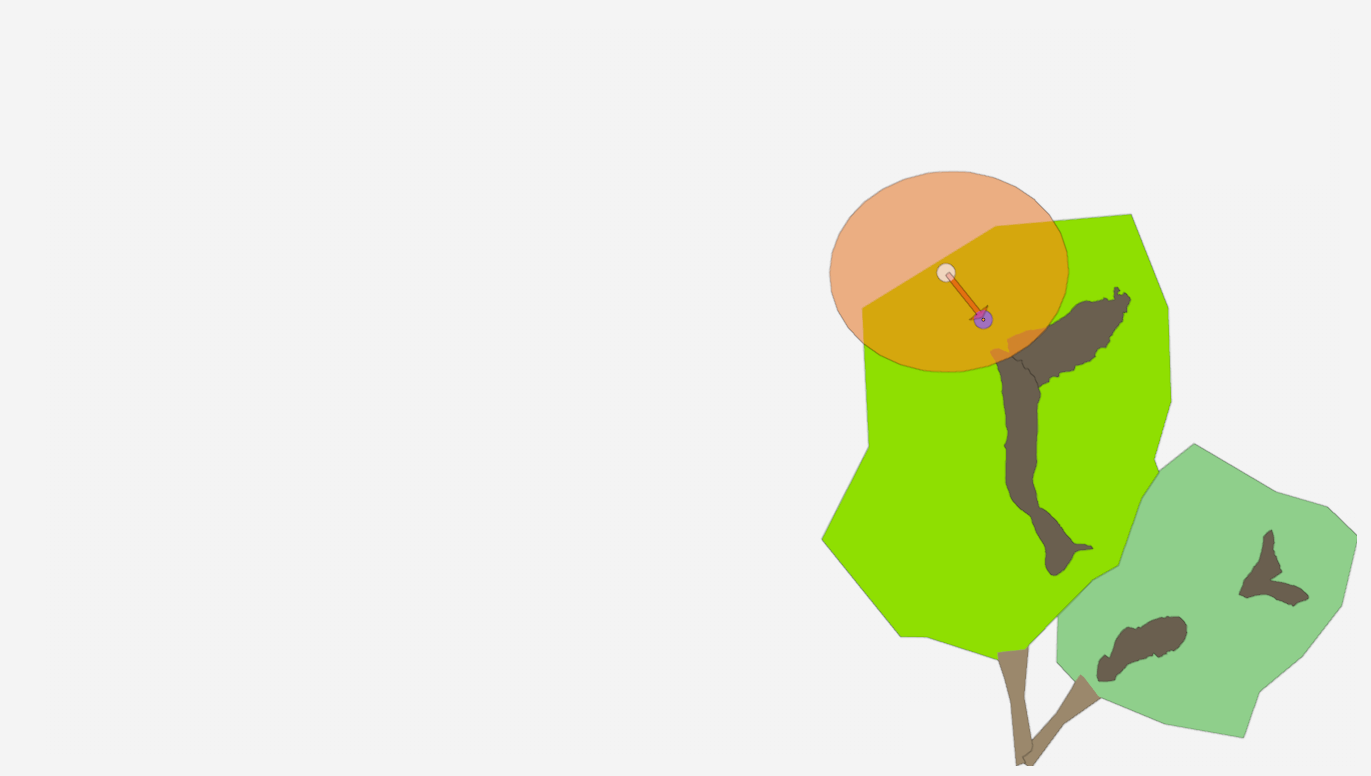}
  \label{subfig:fig2}
  \vspace{+0.1cm}
\end{subfigure}

\caption{Field of view from the camera (left) and Image (right). The figure shows: fruit centroid (white circle) $(\hat{u}_c, \hat{v}_c)$, occlusion centroid (blue circle) $(\bar{u}_c, \bar{v}_c)$ , view frustum (blue cone), its center line $\mathcal{F}^z_{fruit}$ (green arrow) and view gradient $\vec{\mathbf{O}}_f$ (red arrow).}
\label{fig:occlusionid}
\end{figure}

The fruit frame $\mathcal{F}_{fruit}$ is defined as shown in Figure \ref{fig:occlusionid} with the $z$ axis in the direction of the fruit:
\begin{equation} 
\begin{array}{rccc}
\mathbf{f}:  & \mathcal{F}_{cam}  & \rightarrow & \mathcal{F}_{fruit} \vspace{.2cm}\\
& \left( 
\begin{array}{c}
x\\
y\\
z
\end{array}
\right) & \mapsto & 
\mathbf{R}^{\mathbf{p}_{c}}\left( 
\begin{array}{c}
x^\dagger\\
y^\dagger\\
z^\dagger
\end{array}
\right)
\end{array}
\end{equation}

with $\mathbf{R}^{\mathbf{p}_{c}}$ the rotation matrix aligning $\mathcal{F}^{z}_{cam}$ to $\mathbf{p}_{c}$, which can be calculated as in (\ref{eq:rodriguez}).

Under \textit{Hypothesis 2} defined in Section \ref{sec:problem}, we compute the equation of the view frustum which connects camera and fruit, shown in Figure ~\ref{fig:occlusionid} and defined as:
%\begin{equation} 
%\label{eq:FOV}
%\mathrm{FOV}_{xy} = 2 \left( 
%\begin{array}{c}
%\mathrm{arctan}\left(\frac{w}{2f_x}\right)\\
%\mathrm{arctan}\left(\frac{h}{2f_y}\right)\\
%\end{array}
%\right)
%\end{equation}
\begin{equation}
    \frac{{x^\dagger}^2}{{\mathbf{m}({w_f})^2}}+ \frac{{y^\dagger}^2}{{\mathbf{m}({h_f})^2}} = \frac{{{z^\dagger}^2}}{{||\mathbf{p}_{c}||^2}}, ~    \mathrm{s.t.}~~ |z^\dagger| \leq ||\mathbf{p}_{c}||
\end{equation}
%with $r_x$ and $r_y$ as the semi-axis of the ellipsoid along the $x$ and $y$ axis. 
with ${\mathbf{m}({w_f})^2}$ and ${\mathbf{m}({h_f})^2}$ as the width and the height of the fruit in meters.

% We define two hypothesis to simplify the view cone definition:
% \begin{enumerate}
%     \item[] \textit{Hypothesis 2}: Fruits are small compared to the image. $w_f \ll w$ and $h_f \ll h$. 
%     \item[] \textit{Hypothesis 3:} Fruit width $w_f$ and height $h_f$ are similar $w_f \approx h_f \approx r_f$ with $r_f$ as the approximate radius of the fruit.  
% \end{enumerate}

% Under \textit{Hypothesis 2} and \textit{Hypothesis 3}, we can simplify the view cone to a view cylinder:

% \begin{equation}
% \begin{aligned}
%     % {x}_{closest}^2+{y}_{closest}^2 = r^{2},~~ |z| \leq \mathrm{dist}_{cam}(\mathbf{P}_{closest})
% {x^\dagger}^2+{y^\dagger}^2 = r_{f}^{2}\\
% \mathrm{s.t.}~~ |z^\dagger| \leq ||\mathbf{p}_{c}||
% \end{aligned}
% \label{eq:viewcylinder}
% \end{equation}

% with $r_{f}=\max\left[(\mathbf{f}(w_f),\mathbf{f}(h_f)\right]$, the radius of the view cylinder. The center line $\mathbf{l}_{view}$ of the view cylinder is given by:

% \begin{equation}
% \begin{array}{l}
% x^\dagger=0\\
% y^\dagger=0\\
% % z^\prime=z^\prime,~~ |z^\prime| \leq \mathrm{dist}_{cam}(\mathbf{P}_s)
% z^\dagger=z^\dagger : |z^\dagger| \leq ||\mathbf{p}_{c}||
% \end{array}
% \end{equation}
% View cylinder is shown in Figure \ref{fig:occlusionid}. The effect of the \textit{Hypothesis 3} which turns the base of the view cylinder slightly bigger than the fruit can be seen in Figure \ref{fig:cylinder}.

The branch to push is selected among the lines that are within a distance $d_V$ to the view frustum $V$ and have a depth ($z_{branch}$) smaller than the fruit centroid depth ($z_{c}$). These points form $\mathbf{l}^{view}_{branch}$. Then, we project each line $\mathbf{l}_{3D}$ in $\mathbf{l}^{view}_{branch}$ to the image plane $\mathcal{I} \times \mathcal{R}^+$, via $\mathbf{m}^{-1}$. We compute the angle between $\mathbf{I}_{2D}$ and the view gradient $\vec{\mathbf{O}_f}$ as $\gamma = \alpha - \frac{\pi}{2} $; via $\gamma$, to identify the branch which is closer to the perpendicular to $\vec{\mathbf{O}_f}$. The line to push $\mathbf{l}_{push}$ is selected as the $\mathbf{l}_{3D}$ with the smallest $\gamma$. The pushing line decision algorithm is summarized in Algorithm \ref{alg:line_decision}. An example of Hough line transform output $\mathbf{l}^{view}_{branch}$ and the selected pushing line $\mathbf{l}_{push}$ are shown in Figure \ref{fig:hough}.

% \begin{equation}
% \label{eq:viewcylinder}
% \mathbf{I}_{push} = \arg\min_{\mathbf{l} \in \mathbf{l}_{branch}}\left|arccos\left[ \mathbf{m^{-1}}\left({\mathbf{l}}\right)\cdot \vec{\mathbf{O}_f}\right]-\frac{\pi}{2}\right|
% \end{equation}

\begin{algorithm}

 \caption{Pushing line decision algorithm}
 \begin{algorithmic}[1]
 \renewcommand{\algorithmicrequire}{\textbf{Input:}}
 \renewcommand{\algorithmicensure}{\textbf{Output:}}
 \REQUIRE Line segments $\mathbf{l}_{branch}$, 
            Occlusion direction $\vec{\mathbf{O}_f}$
 \ENSURE Line to push $\mathbf{l}_{push}$
 \STATE Filter branch segments: 
 $\mathbf{l}^{view}_{branch} \leftarrow  \mathrm{dist}(\mathbf{l}_{branch}, V) <  d_V$ and $z_{branch} < z_{c}$
    \FOR{$\mathbf{l_{3D}} \in \mathbf{l}^{view}_{branch}$ }
        \STATE Project the $\mathbf{l}$ into image plane: $\mathbf{l}_{2D} \leftarrow \mathbf{m^{-1}}\left({\mathbf{l_{3D}}}\right)$
        \STATE Get the angle $(\mathbf{l}_{2D}, \vec{\mathbf{O}_f})$ : $\alpha \leftarrow \arccos\left[ \mathbf{l}_{2D} \cdot \vec{\mathbf{O}_f}\right]$
        \STATE Check $\alpha$ is how far from $\frac{\pi}{2}$: 
         $\gamma \leftarrow \alpha - \frac{\pi}{2} $
    \ENDFOR
    \RETURN $\mathbf{l}_{push} \leftarrow \mathbf{l_{3D}}$ with smallest $\gamma$
 \end{algorithmic}
 \label{alg:line_decision}
 \end{algorithm}

\subsection{Push Action}

If the pushing line $\mathbf{l}_{push}$ intersects with the view frustum, we select the initial pushing point among the points that intersect with the circumference of the view frustum $V_C$, opting for the one closer to the robot  to minimize robot self-occlusions. Otherwise, we select the closest point to the view frustum, belonging to $\mathbf{l}_{push}$. We define the initial pushing point $\mathbf{p}_{push}$ as:

\begin{equation}
\mathbf{p}_{push} = 
\begin{cases*}   \mathbf{p} \cap V_C {\text{~close to robot}} 
& {\text{if}} $\mathbf{l}_{push} 
\cap V = \emptyset$ 
\\
  \arg\min \mathrm{dist}(\mathbf{p}, V)   & {\text otherwise}
\end{cases*}
\end{equation}
with $\mathbf{p} \in \mathbf{l}_{push}$ and $\mathrm{dist}$ as the distance in 3D space and $V$ the view frustum. 
Push magnitude is set to the largest distance between points in $\mathbf{M}^\prime$:

\begin{equation}
 d_{push}= \max_{\mathbf{p}_i, \mathbf{p}_j \in \mathbf{M}^\prime} \left\| \mathbf{p}_i - \mathbf{p}_j \right\|
\end{equation}

Finally, we move the robot end-effector from point $\mathbf{p}_{push}$ on a straight segment of length $d_{push}$, in the view gradient direction, $\vec{\mathbf{O}}_f $. The end-effector orientation is set to maintain a $\pi/4$ angle wrt the camera principal axis, and directed away from the image plane to prevent robot self-occlusions. 

 \begin{figure}[t]
	\centering {\centering\includegraphics[width=0.99\columnwidth]{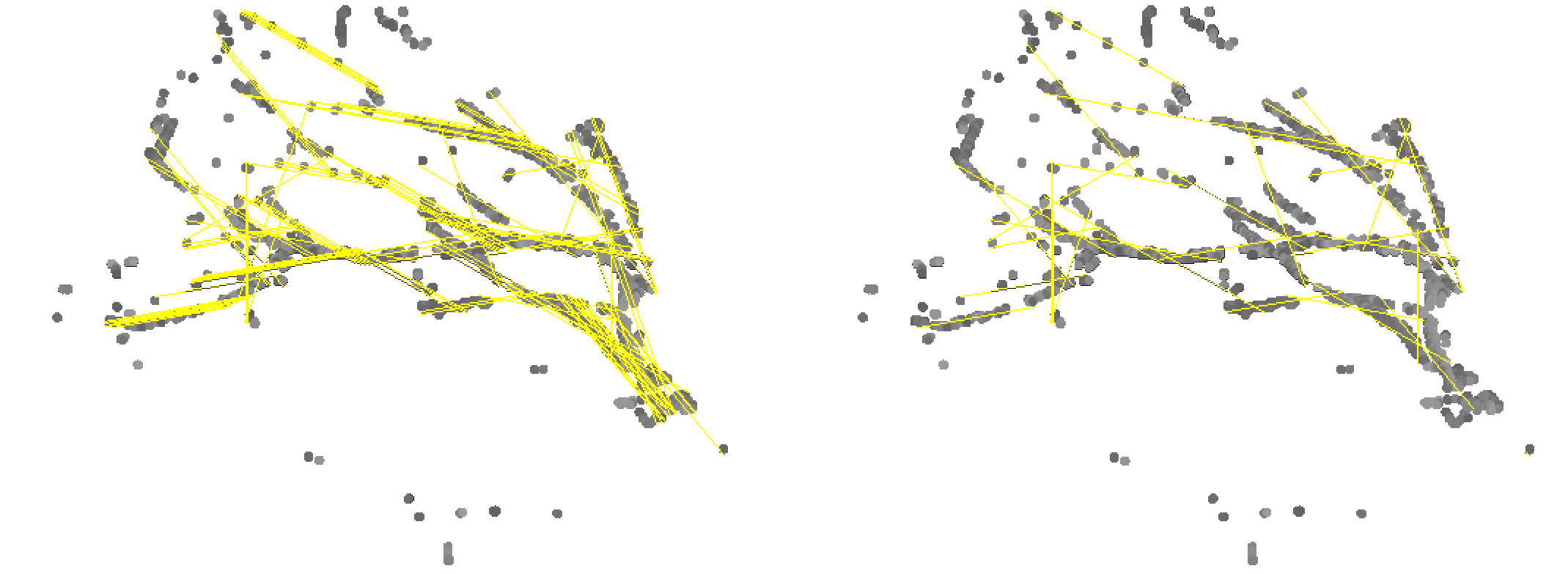}}
    \caption{Line detection before (left) and after (right) removing similar lines}

    \label{fig:unfiltered} 
\end{figure}

\begin{figure}[t]
\centering
\begin{subfigure}{0.99\columnwidth}
  \centering
  \includegraphics[width=0.99\columnwidth]{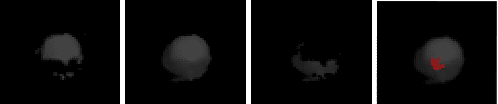}
  \vspace{0.15cm}
  \label{subfig:fig1}
\end{subfigure}
\vspace{0.15cm}
\begin{subfigure}{0.99\columnwidth}
  \centering
  \includegraphics[width=0.99\columnwidth]{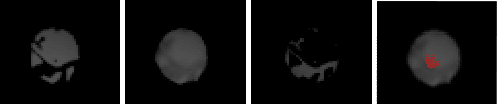}
  \label{subfig:fig2}
\end{subfigure}
\vspace{0.15cm}
\begin{subfigure}{0.99\columnwidth}
  \centering
  \includegraphics[width=0.99\columnwidth]{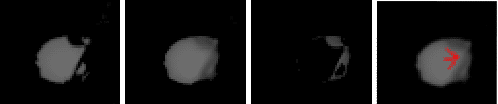}
  \label{subfig:fig3}
\end{subfigure}
\begin{subfigure}{0.99\columnwidth}
  \centering
  \includegraphics[width=0.99\columnwidth]{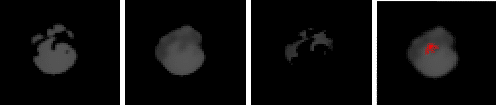}
  \label{subfig:fig4}
\end{subfigure}
\caption{From left to right: segmented fruit in depth image input to the fruit estimator, estimation result, difference between input and output, and view gradient (red vector) on the estimated fruit.}
\label{fig:view_grad}
\end{figure}

\section{EXPERIMENTS} \label{sec:experiment}

Four experiments are conducted in three scenes under different light conditions and fruits (i.e. apple, lemon, orange) to test the perception algorithms. We run a case study with real robot experiment on an artificial orange tree.  The video of the experiments attached to the paper is also available online at: \href{youtu.be/watch?v=hGmj7sBb9vw}{youtu.be/watch?v=hGmj7sBb9vw}.

We use the right arm (7-DOF KUKA LWR) of BAZAR robot \cite{cherubini2019collaborative} and an Intel Realsense D-435 RGB-D camera. A metal rod is attached at the center of the last arm link as a tool to perform the push action. We use Intel Realsense SDK for camera acquisition, and OpenCV for image processing. The robot is controlled using ROS Noetic with MoveIt.

\begin{figure}[t]
	\centering {\centering\includegraphics[width=1.0\columnwidth]{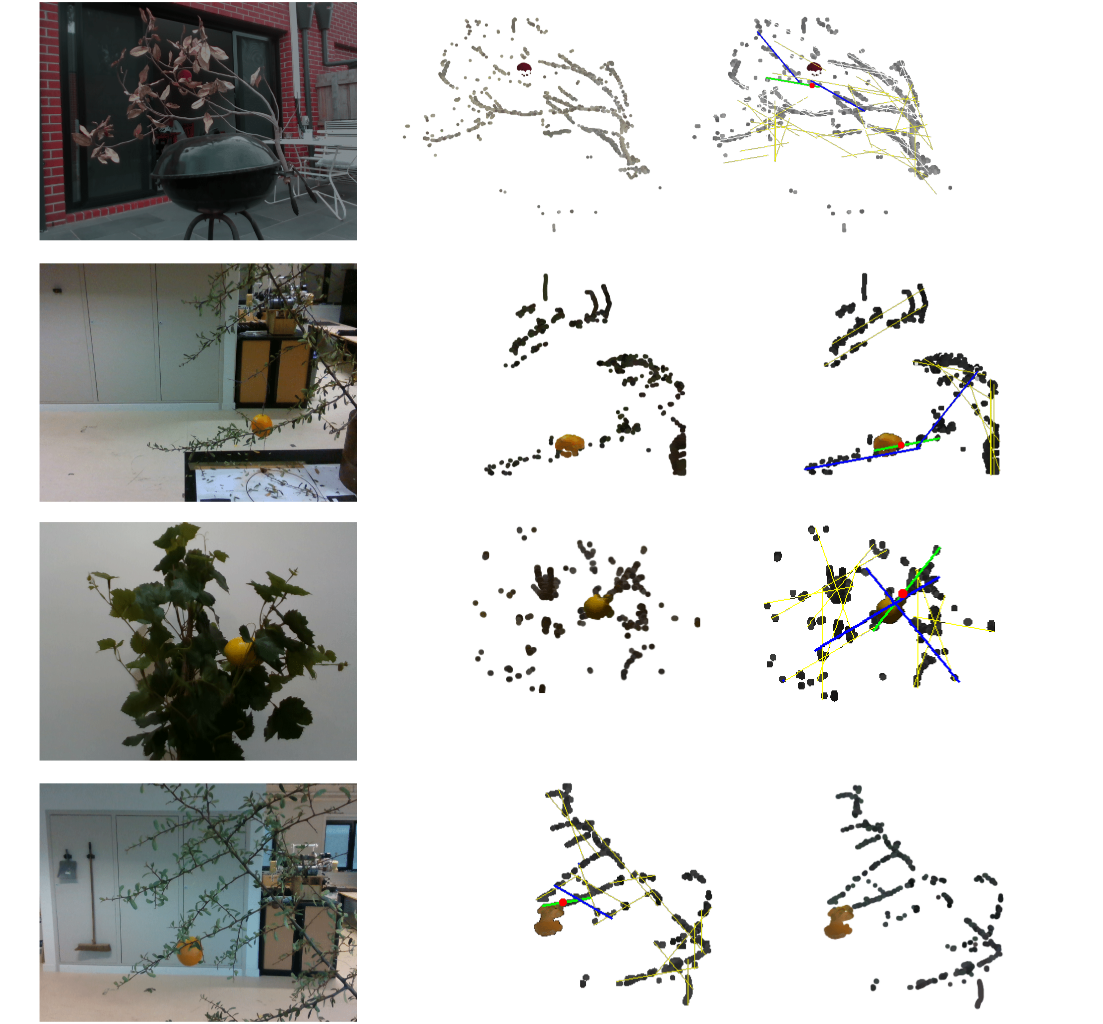}}
	\caption{From left to right: scene, segmentation results, and line detection results. For line detection: detected and filtered lines are yellow, lines below the distance threshold are blue, and the selected line is green. The pushing point is red.}
    \label{fig:lines} 
\end{figure}

\begin{figure}[t]
	\centering 
	\includegraphics[width=1.0\columnwidth]{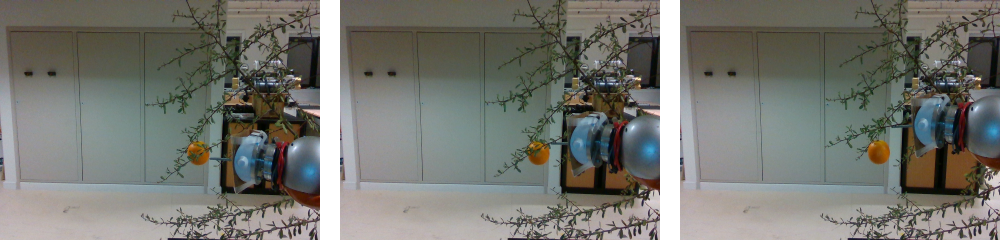}
	\caption{Robot branch pushing demonstration.}
 \label{fig:robot_push}
\end{figure}

The results of fruit appearance estimation, with the difference between the segmented fruit $\mathbf{M}'_{M}$ and its estimation $\mathbf{\bar{D}}_M^\prime$ as well as the view gradient $\vec{\mathbf{{O}}}_f$ are shown in Figure \ref{fig:view_grad}. Fruits used in these experiments are: apple, orange, lemon, orange from top to bottom. The results suggest that while the shape of the fruit is accurately predicted, the depth estimation is less reliable. However, this is not crucial since we will binarize the image to obtain $\mathbf{\bar{M}}^\prime$ from these predictions.

Different scenes are shown in Figure \ref{fig:lines} with the branch and fruit segmentation results, including detected branch lines $\mathbf{l}_{branch}$, pushing line $\mathbf{l}_{push}$ and pushing point $\mathbf{p}_{push}$. The outcomes indicate that branches are generally accurately predicted, although there are instances of non-existent branch lines being detected. The selected pushing lines $\mathbf{l}_{push}$ and pushing points $\mathbf{p}_{push}$ appear reasonable from an empirical standpoint.

Figure~\ref{fig:unfiltered} compares branch segmentation, before and after removing similar lines via Algorithm \ref{alg:line_find}. Parameters used in this example: $\rho_{th} = 25$ and $\theta_{th} = \frac{\pi}{6}$ rad. In other examples, $\rho_{th}$ values vary between $15$ - $25$. The last column of Figure \ref{fig:lines} is used as a case study to clear the occlusion with the robot. The robot in action is shown in Figure \ref{fig:robot_push}. The robot successfully clears the occlusion.

\section{CONCLUSION}\label{sec:conclusion}

This paper presents a novel method for clearing fruit occlusions in agricultural applications. We begin by segmenting branches and fruit in the scene. We use a deep generative model to estimate the fruit's whole appearance. Then, we introduce a 3D extension of the 2D Hough transform, to detect branches as straight line segments. This enables us to identify the branch responsible for occlusion. We use the robot, to push the identified branch and clear the occlusion. We demonstrate the effectiveness of our perception methods under three different lighting conditions and across different types of elliptical fruit i.e. apple, orange, lemon. We present a case study involving a real robot demonstration.

While effective in controlled cases, our method faces challenges with branch segmentation and line detection, especially in complex scenarios and multiple occlusions. Scaling to multiple trees or integrating multiple robotic arms requires more research and robust real-time processing. Scalability involves adapting to different orchard types, tree densities, and environmental conditions, with customization needed for tree heights and shapes, and integration with existing machinery. Sensor fusion and adaptive machine learning can increase performance across varied settings.

Our approach of addressing occlusions by physically pushing obstacles can enhance detection precision by establishing ground truth instead of solely relying on estimations. In the future, this method can be generalized to different fruit shapes and integrate with multi-arm robots to create paths for fruit accessibility. We believe that the methods presented in this paper represent a step forward in agricultural robotics.

\addtolength{\textheight}{-12cm}   % This command serves to balance the column lengths

%%%%%%%%%%%%%%%%%%%%%%%%%%%%%%%%%%%%%%%%%%%%%%%%%%%%%%%%%%%%%%%%%%%%%%%%%%%%%%%%

%%%%%%%%%%%%%%%%%%%%%%%%%%%%%%%%%%%%%%%%%%%%%%%%%%%%%%%%%%%%%%%%%%%%%%%%%%%%%%%%

%%%%%%%%%%%%%%%%%%%%%%%%%%%%%%%%%%%%%%%%%%%%%%%%%%%%%%%%%%%%%%%%%%%%%%%%%%%%%%%%

\bibliographystyle{IEEEtran}
\bibliography{IEEEabrv,ref}

\end{document}